\documentclass[10pt,twocolumn,letterpaper]{article}
\usepackage{cvpr}
\usepackage{times}
\usepackage{epsfig}
\usepackage{graphicx}
\usepackage{amsmath}
\usepackage{amssymb}
\usepackage{booktabs}

\usepackage{threeparttable}
\usepackage{float}
\usepackage{verbatim}
\usepackage[dvipsnames]{xcolor}
\usepackage{url}

\usepackage[breaklinks=true,bookmarks=false,colorlinks=true]{hyperref}

\cvprfinalcopy % *** Uncomment this line for the final submission

\newcommand*{\affaddr}[1]{#1} 
\newcommand*{\affmark}[1][*]{\textsuperscript{#1}}
\newcommand*{\email}[1]{ \small \texttt{#1}}
\renewcommand{\thefootnote}{\fnsymbol{footnote}}

\def\cvprPaperID{6565} % *** Enter the CVPR Paper ID here

\begin{document}

%%%%%%%%% TITLE
\title{State-Aware Tracker for Real-Time Video Object Segmentation}

%\title{Enhancing Video Object Segmentation by Tracking}
%\setcounter{footnote}{-1}
\author{%
Xi Chen\affmark[1]\footnotemark[2], Zuoxin Li\affmark[2], Ye Yuan\affmark[2], Gang Yu\affmark[2], Jianxin Shen\affmark[1], Donglian Qi\affmark[1]\\
\affaddr{\affmark[1] College of Electrical Engineering, Zhejiang University}, 
\affaddr{\affmark[2]Megvii Inc. }\\
\email{\{xichen\_zju,J\_X\_Shen,qidl\}@zju.edu.cn},\\
\email{\{lizuoxin,yuanye,yugang\}@megvii.com}\\
%\affaddr{\LaTeX\ University}%
}

\maketitle

\renewcommand{\thefootnote}{\fnsymbol{footnote}}
\footnotetext[2]{This work was done during an internship at Megvii Inc.}

%%%%%%%%% ABSTRACT
\begin{abstract}
In this work, we address the task of semi-supervised video object segmentation~(VOS) and explore how to make efficient use of video property to tackle the challenge of semi-supervision. We propose a novel pipeline called State-Aware Tracker~(SAT), which can produce accurate segmentation results with real-time speed.
For higher efficiency,  SAT takes advantage of the inter-frame consistency and deals with each target object as a tracklet. 
For more stable and robust performance over video sequences, SAT gets awareness for each state and makes self-adaptation via two feedback loops. One loop assists SAT in generating more stable tracklets. The other loop helps to construct a more robust and holistic target representation. SAT achieves a promising result of 72.3\% $\mathcal{J \& F}$ mean with 39 FPS on DAVIS2017-Val dataset, which shows a decent trade-off between efficiency and accuracy. Code will be released at \url{github.com/MegviiDetection/video\_analyst}.
\end{abstract}

%%%%%%%%% BODY TEXT
\section{Introduction}

Semi-supervised video object segmentation~(VOS) requires to segment target objects over video sequences with only the initial mask given, which is a fundamental task for computer vision.  In VOS task, the initial mask is provided as visual guidance.  Nevertheless, throughout a video sequence, the target object can undergo large pose, scale, and appearance changes. Moreover, it can even meet abnormal states like occlusion, fast motion, and truncation. Therefore, it is a challenging task to make a robust representation over video sequences in a semi-supervised manner.

Luckily, video sequence brings additional context information for VOS task. First, the inter-frame consistency of video makes it possible to pass information efficiently between frames. Furthermore, in VOS tasks, information from preceding frames could be regarded as the temporal context, which can provide helpful cues for the following predictions. Hence, making efficient use of the additional information brought by video is of great importance for VOS tasks. 

\begin{figure}[t]
\centering
\includegraphics[width=\linewidth]{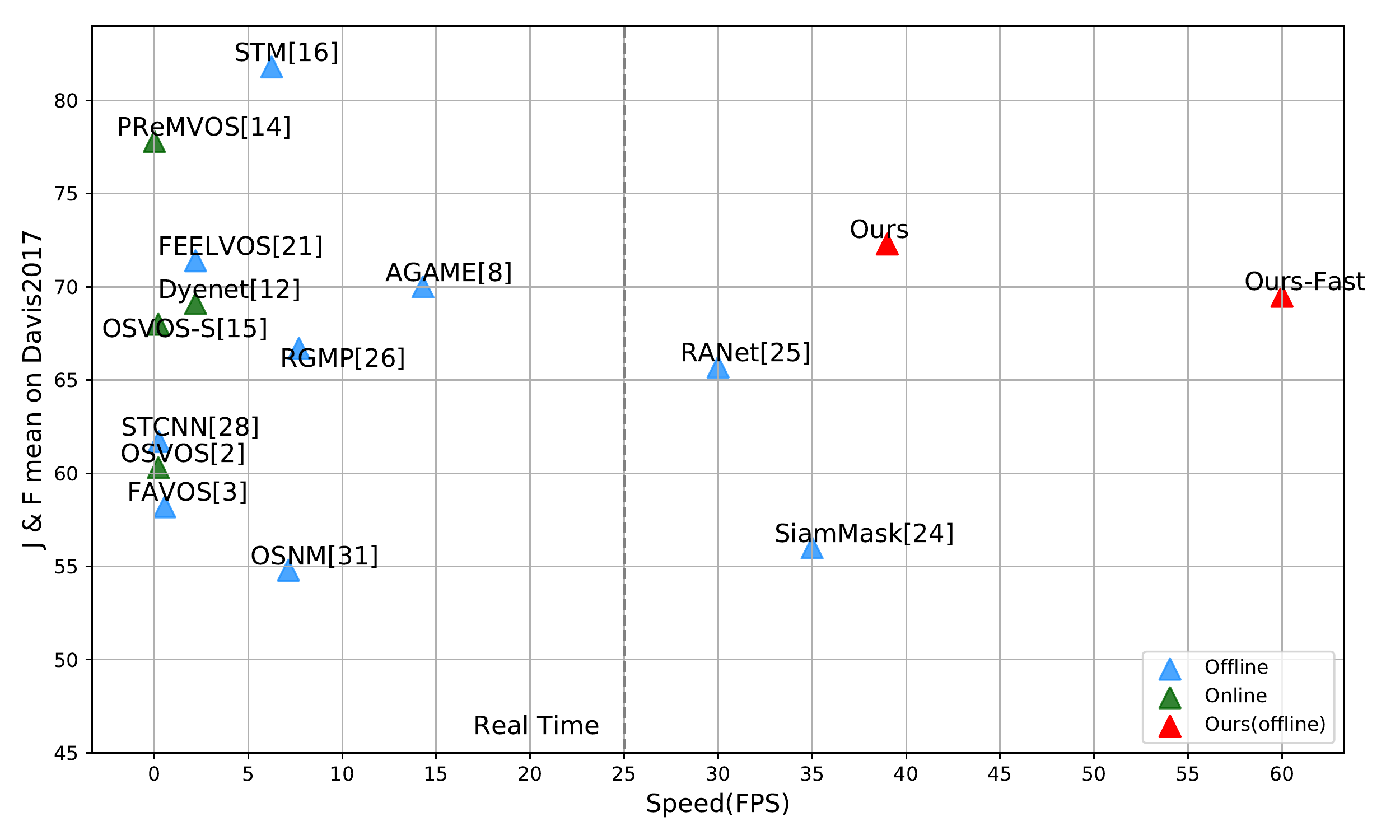}
\caption{ Accuracy versus  speed  on DAVIS2017-Val dataset. Some previous methods achieve high accuracy with slow running speed. Others sacrifice too much accuracy for the faster speed. Our method achieves a decent speed-accuracy trade-off.    }
\label{fig:pic1}
\end{figure}

However, previous works do not make good use of the characteristics of videos. \cite{osvos,osvos-s,onavos,RGMP,dyenet} completely ignore the relation between frames and deal with each frame independently, which causes tremendous information waste. Other methods~\cite{voigtlaender2019feelvos,masktrack,osnm,xiao2018monet,osnm} use feature concatenation, correlation, or optical flow to propagate predicted mask or feature from the previous frame to the current frame, but they have apparent drawbacks. First, previous works usually propagate information on full images, while the target object usually occupies a small region. In this case, operations on full images can cause redundant computation. Furthermore, the target object can undergo different states throughout the video, but these methods apply fixed propagation strategies without adaptation, which makes them unstable over long sequences. Moreover, they only seek cues from the first or the previous frame for target modeling, which is not enough for a holistic representation.  As a result, most existing methods can not tackle VOS with both satisfactory accuracy and fast speed.  Therefore, a more efficient and robust pipeline for semi-supervised video object segmentation is required. 

In this paper,  we reformulate VOS as a continuous process of state estimation and target modeling, in which segmentation is a specific aspect of state estimation.  Specifically, we propose a simple and efficient pipeline called State-Aware Tracker~(SAT). Taking advantage of the inter-frame consistency, SAT takes each target object as a tracklet, which not only makes the pipeline more efficient but also filters distractors to facilitate target modeling. In order to construct a more reliable information flow, we propose an estimation-feedback mechanism that enables our model to be aware of the current state and make self-adaptation for different states. For a more holistic target modeling,  SAT uses the temporal context to construct a global representation dynamically to provide robust visual guidance throughout the video sequence. As demonstrated in Fig.~\ref{fig:pic1}, SAT achieves competitive accuracy and runs faster than all other approaches on DAVIS2017-Val dataset.

\begin{figure}[t]
\centering
\includegraphics[width=\linewidth]{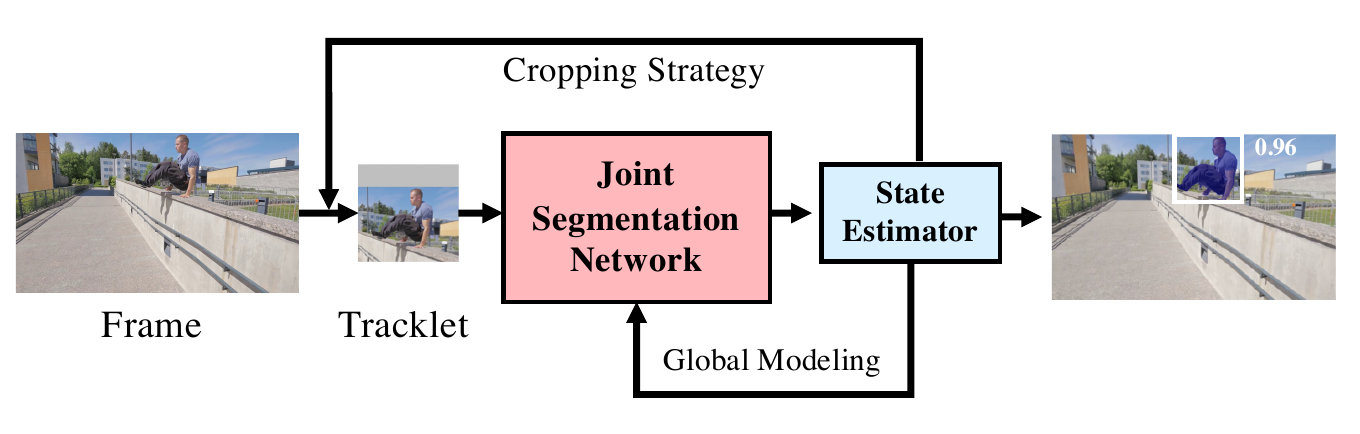}
\caption{A simplified demonstration of our video object segmentation pipeline.}
\label{fig:pipe_small}
\end{figure}

A simplified illustration of our pipeline is provided in Fig.~\ref{fig:pipe_small}. The inference procedure could be summarized as \textit{Segmentation - Estimation - Feedback}. First, SAT crops a search region around the target object and takes each target as a tracklet. Joint Segmentation Network predicts masks for each tracklet. Second, State Estimator evaluates the segmentation result and produces a state score to represent the current state. Third, based on state estimation results, we design two feedback loops. Cropping Strategy Loop picks different methods adaptively to predict a bounding box for the target. Then, we crop the search region for the next frame according to the predicted box. This switching strategy makes the tracking process more stable over time. Meanwhile, Global Modeling Loop uses the state estimation results to update a global feature dynamically. In return, the global feature can assist Joint Segmentation Network in generating better segmentation results.

To verify the effectiveness of our method, we conduct extensive experiments and ablation studies on DAVIS2016, DAVIS2017 and YouTube-VOS datasets. Results show that SAT achieves strong performance with a decent speed-accuracy trade-off. 
Our main contributions can be summarized as follows: 
(1) We re-analyze the task of semi-supervised video object segmentation and develop State-Aware Tracker, which reaches both high accuracy and fast running speed on DAVIS benchmarks. 
(2) We propose a state estimation-feedback mechanism to make the VOS process more stable and robust over time.
(3) We propose a new method of constructing global representation for the target object to provide more robust guidance.  

%-------------------------------------------------------------------------
\begin{figure*}[h]
\centering
\includegraphics[width=\linewidth]{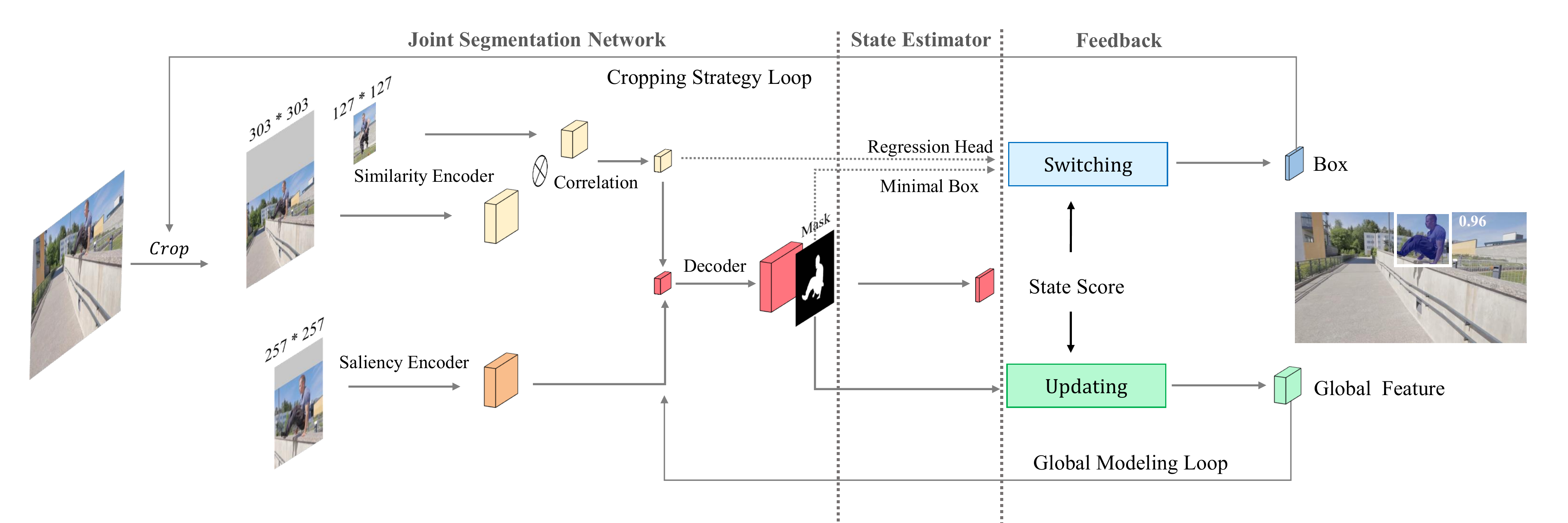}
\caption{An overview of our video object segmentation pipeline. SAT can be divided into three parts by the dotted line in {\color{gray}gray}: Joint Segmentation Network, State Estimator, and Feedback. \textbf{Joint Segmentation Network} fuses the feature of the saliency encoder~(in {\color{orange}orange}), the similarity encoder~(in {\color{yellow}yellow}), and the global feature~(in {\color{green}green}), and then decodes the fused feature to predict a mask. Afterward, \textbf{State Estimator} evaluates the prediction result and calculates a state score to represent the current state. Finally,  based on the state estimation result,  \textbf{Cropping Strategy Loop} switches the cropping strategy to keep a more stable tracklet. \textbf{Global Modeling Loop} constructs a global representation to enhance the feature of the segmentation network.  
}
\label{fig:pipeline_big}
\end{figure*}

\section{Related Works}
Video object segmentation task aims at segmenting target object in video frames given the initial mask of the first frame. In recent years, a wide variety of methods has been proposed to address this challenge.
\textbf{Online learning based methods }:  In order to distinguish the target object from background and distractors, online-learning based methods fine-tune the segmentation network on the first frame.  
OSVOS~\cite{osvos} fine-tunes a pretrained segmentation network on the first frame of test videos. 
OnAVOS~\cite{onavos} extends OSVOS by developing an online adaptation method.
OSVOS-S~\cite{osvos-s} introduces instance information to enhance the performance of OSVOS. Lucid tracker~\cite{lucid} studies the data augmentation method for the first frame of test videos and brings significant improvement. 
%DOLF~\cite{dolf} designs a target appearance model to apply online fine-tune more efficiently. 
Many other methods~\cite{wang2019ranet,luiten2018premvos,PTS} take online learning as a boosting trick to reach better accuracy. Online learning has been proved to be an effective way to make VOS models more discriminative for the target object. However, it is too computational expensive to be used in practical applications. Generally, online models address the challenge of semi-supervised learning via updating model weight, which entails extensive iterations of optimization. Instead of updating model weight, our method updates a global representation via dynamic feature fusion, which tackles the challenge of target modeling more efficiently.

\textbf{Offline learning based methods } : Offline methods exploit the use of the initial frame and pass target information to the following frames via propagation or matching. MaskTrack~\cite{masktrack} concatenates the predicted mask of the previous frame with the image of the current frame  to provide spatial guidance.  
%OSMN~\cite{osnm} uses a visual modulator to extract visual guidance from the initial frame and constructs a spatial modulator to propagate location prior to adjacent frames.
FEELVOS~\cite{voigtlaender2019feelvos}  develops pixel-wise correlation to pass location-sensitive embeddings over consecutive frames. RGMP~\cite{RGMP} uses a siamese encoder to capture local similarities between the search image and the reference image. AGAME~\cite{agame} proposes a probabilistic generative model to predict target and background feature distributions. These methods do not entail computational expensive online fine-tuning, but they still cannot reach fast speed due to inefficient information flow.
Moreover, They usually suffer sub-optimal accuracy because they lack robust target representation. Our method is also offline trained and propagates visual cues from frame to frame. Different from previous, we take each object as a tracklet and apply self-adaptation, thus making the information flow more efficient and stable. Besides, we use the temporal context to update a global representation, which provides more robust guidance over video sequences.

\textbf{Tracking based methods}: FAVOS~\cite{FAVOS} develops a part-based tracking method to track local regions of the target object. SiamMask~\cite{siammaks} narrows the gap between object tracking and object segmentation by adding a mask branch on SiamRPN~\cite{siamrpn}, and it runs much faster than previous works. %cite{renrobust,boltvos} split video object segmentation into two sub-tasks: box-tracking and segmentation, and they perform competitive results on DAVIS benchmark. 
These tracking-based methods take tracking and segmentation as two separated parts. The segmentation result is not involved in the process of tracking, and it could be regarded as post-processing for the tracker. Different from previous works, we fuse object tracking and segmentation into a truly unified pipeline, in which there is no restrict boundary between tracking and segmentation. In our framework, these two tasks cooperate closely and enhance each other.  \\

\begin{comment}
\textbf{Real-time segmentation}: Inference speed is important for segmentation tasks, especially when it comes to practical applications. Some works deal with this challenge by designing a multi-branch structure.  A classical approach is decoupling the high-level feature and low-level into separated branches and design each branch specifically. BiseNet ~\cite{bisenet} proposes a bilateral architecture with a spatial path and a context path. ICNet~\cite{ICnet} designs a system with three branches and takes cascade image inputs. Image with high resolution passes a shallow extractor to provide spatial information while low-resolution image passes a deeper network to extract semantic information. Inspired by those works, we propose a multi-branch structure to decouple the functions of each branch, and we set different input resolutions for each branch according to its function. In this way, each branch of our network has its own mission, and they cooperate with each other efficiently.\\
\end{comment}

\section{Method}

\subsection{Network Overview}
\begin{comment}
In this work, we propose a novel pipeline called State-Aware Tracker~(SAT). Taking advantage of the inter-frame consistency, our pipeline deals with each target object as a tracklet. To make our pipeline more stable and robust over time,  we propose a state estimation-feedback mechanism and design two feedback loops. One loop switches cropping strategy to generate a more stable tracklet. The other constructs a global representation to enhance the feature of the target object. In general, SAT has awareness for each state, and it can make self-adaptation and self-enhancement according to different states.
\end{comment}
In this work, we propose a novel pipeline called State-Aware Tracker~(SAT), which gets high efficiency via dealing with each target as a tracklet. Besides, SAT gets awareness for each states and develop self-adaptation via two feedback loops. % This mechanism enables it to get stable and robust performance over video sequences.   

As in Figure \ref{fig:pipeline_big}, we describe our inference procedure with three steps : \textbf{Segmentation - Estimation - Feedback.} 
First, \textit{Joint Segmentation Network}  fuses the feature of the similarity encoder, the saliency encoder, and the global feature to produce a mask prediction. Second, \textit{State Estimator} evaluates the segmentation result and describes the current state with a state score and estimates whether it is a normal state or an abnormal state. Third, we construct two feedback loops to make self-adaptation for different states. In  \textit{Cropping Strategy Loop}, if it is a normal state, we use the predicted mask to generate a minimal bounding box. Otherwise, we use a regression head to predict the bounding-box and apply temporal smoothness. Then, based on the predicted box, we crop the search region for the next frame.
In \textit{Global Modeling Loop}, we use the state estimation results, the predicted mask and the current frame image patch to update a global feature, and use the global feature to enhance  \textit{Joint Segmentation Network} for better segmentation results.  
In the following section, we introduce each stage in detail. %More details refer to our source code which will be made available upon publication. \\

\subsection{Segmentation}
As shown in Figure \ref{fig:pipeline_big}, the branch on bottom denotes the saliency encoder, and the two branches on top demonstrate the similarity encoder. For the input of the saliency encoder, we crop a relatively small region around the target to filter distractors, and we zoom it to a larger resolution to provide more details. In this way, the saliency encoder can extract a clean feature with rich details for the salient object of the input image patch. In this work, we use a shrinked ResNet-50~\cite{resnet} for the saliency encoder.

The similarity encoder takes a larger search region of the current frame and a target region of the initial frame as input.  It uses feature correlation to encode appearance similarities between the current image and the target object. This \textbf{correlated feature} provides appropriate supplementary for the saliency encoder to distinguish the target object from distractors. 
 In this work, the implementation of the similarity encoder follows SiamFC++~\cite{siamfc++} with Alexnet~\cite{alex} backbone.
 
%The saliency encoder takes an image of $257\times257$ as input, and the similarity encoder takes a larger one of $303\times303$. Considering the similarity network does not exert padding for convolution operation, the output feature of the siamese encoder has the same size with the  4-stage feature of the saliency encoder.
%We use element-wise addition to fuse those two feature maps to generate a more distinguishing representation. We also fuse the global feature produced by the Global Modeling Loop.
The saliency encoder extracts a class-agnostic feature for the target object, which is clean but lacks discrimination. Meanwhile, the correlated feature of the similarity encoder provides instance-level appearance similarity, which assists our network to distinguish the target object from distractors. In addition, the global feature updated by the Global Modeling Loop provides a holistic view for the target object, which is robust for visual variants over long sequences. In Joint Segmentation Network, we fuse these three features via element-wise addition to obtain a strong high-level feature with both discrimination and robustness.  

After the feature fusion, we upsample the high-level feature by bilinear interpolation and concatenate it with low-level features of the saliency encoder successively. 
Consider that the input image of the saliency encoder is cropped around the target with high resolution, the low-level feature of the saliency encoder is clean and full of details, which assists the Joint Segmentation Network to decode a high-quality mask with fine contours.   \\

\subsection{Estimation}
During the process of video segmentation, the target object can go through various states, such as well-presented, truncated, occluded, even can run out of the search region. In different states, we should take different actions to crop the search region for the next frame and apply different strategies to update the global representation.

State Estimator evaluates each local state with a state score and divides all states into two categories:  normal state and abnormal state. We analyze that the state of the target object could be described by the mask predicting confidence and the mask concentration. As shown in Tab. \ref{tab:states}, when the target is well-presented in the current image, the mask predicting confidence tends to be high, and the predicted mask is usually spatially concentrated. When the target gets truncated, the predicted mask tends to be separated into several parts, and it leads to low spatial concentration. When the target is occluded or runs out of the search region, the model usually predicts with low confidence.

\begin{table}[H]
\centering 
\small
\tabcolsep=2.3pt
\begin{tabular}{ccccc}
\toprule
& \textit{Confidence} & \textit{Concentration } & \textit{State}  \\
\midrule
Well-Presented  & High 	& High & Normal  \\
%\midrule
Truncated     & - 	& Low & Abnormal  \\
Occluded     & Low 	& -  & Abnormal \\
Disappear     & Low 	& -  & Abnormal \\
%Badly-Segmented & low 	& low  & abnormal \\
\bottomrule
\end{tabular}
\caption{ State estimation criterion. - denotes that the result does not influence the state estimation, which can be either high or low in this case.  }\label{tab:states}%添加标题
\end{table}

Therefore, we propose a confidence score $\mathcal S_{cf}$ to denote the mask predicting confidence, and a concentration score $\mathcal S_{cc}$ to represent the geometric concentration for the predicted mask.
We calculate the confidence score as Eq.~\ref{eq:confidence}, where  $\mathcal P_{i,j}$ denotes mask prediction score at location $(i,j)$, and $\mathcal M$ represents predicted binary mask. $\mathcal M_{i,j}$ equals 1 when the pixel at $(i,j)$ is predicted as foreground, otherwise it equals 0.
\begin{equation}
\mathcal S_{cf} = \frac{ \; \sum_{i,j} \mathcal P_{i,j} \cdot \mathcal M_{i,j} \;  }{ \; \sum_{i,j} \mathcal M_{i,j} \; }\label{eq:confidence}
\end{equation}
We define concentration score as the ratio of the max connected region area to the total area of the predicted binary mask. As in Eq.~\ref{eq:concentration}, $|\mathcal R_i^c|$ denotes the pixel number of the \textit{i th} connected region of the predicted mask.\\
\begin{comment}
\begin{equation}
\mathcal S_{cc} = \frac{\; max(Area_i)\; }{\;\sum_{i}^{n} Area_i \;}\label{eq:concentration}    
\end{equation}
\end{comment}
\begin{equation}
\mathcal S_{cc} = \frac{\; max(\{|\mathcal R_1^c|,|\mathcal R_2^c|, \cdots, |\mathcal R_n^c|\})\; }{\;\sum_{1}^{n} |\mathcal R_i^c| \;}\label{eq:concentration}    
\end{equation}

Finally,  we calculate a state score $\mathcal S_{state}$ as Eq.~\ref{eq:state score}.  If $S_{state} > \mathcal T $, we estimate the current state as a normal state. Otherwise, we judge it as an abnormal state. In this work, we set  $\mathcal T  = 0.85$ according to the result of the grid search.
\begin{equation}
\mathcal S_{state} = \mathcal S_{cf} \times \mathcal S_{cc}\label{eq:state score}    
\end{equation}

\subsection{Feedback}
Based on the estimation result, we construct two feedback loops. One loop switches the cropping strategy to make our tracker more stable over time. The other loop updates a global representation to enhance the process of segmentation. 

\textbf{\textit{Cropping Strategy Loop:}}  For each frame, we generate a bounding box for the target object and crop the search region for the next frame according to the box. In order to maintain a stable and accurate tracklet, we design two box generation strategies and switch the strategy for different states.
For normal states, we select the largest connected region of the binary mask and calculate its minimal bounding box to indicate the position of the target. We use the largest connected region in order to avoid the interference of small pieces of false-positive predictions.
For abnormal states, we add a regression head after the similarity encoder to predict a bounding box, then apply a temporal smoothness on location, scale, and ratio. In this work, we construct our regression head following SiamFC++\cite{siamfc++}.

Considering that mask can provide a more accurate representation for object contours when the object is well-presented, mask-box can predict a more accurate location in normal states. Furthermore, the mask-box corresponds to a smaller search region, which makes it more robust for distractors. In contrast, regression-box is generated from a larger search region, so it can retrieve the object when it runs fast. When the object is truncated, the regression-box can provide complete predictions for the target object. In addition, with the help of the temporal smoothness, the regression-box can still indicate a reasonable location if the object is occluded or even disappeared.

\begin{figure}[H]
\newcommand{\image}{\includegraphics[width=0.32\columnwidth]}
\centering 
\tabcolsep=0.05cm
\renewcommand{\arraystretch}{0.06}
\begin{tabular}{ccc}
\vspace{1mm}
 {\footnotesize Distractor} & {\footnotesize Truncation} & {\footnotesize Fast Motion} \\
\vspace{1mm}
\image{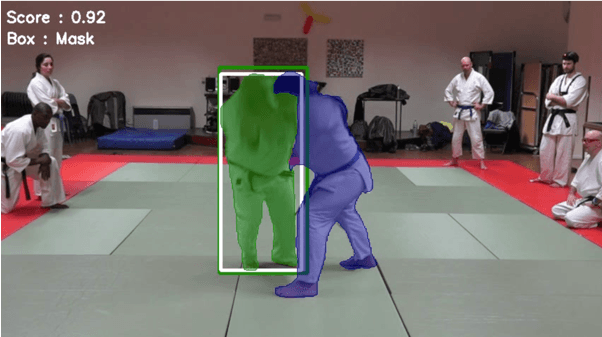} &
\image{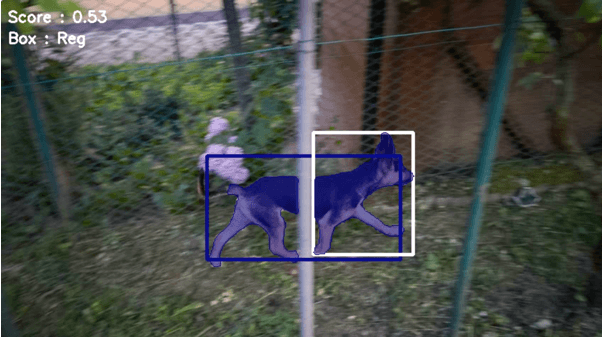} &
\image{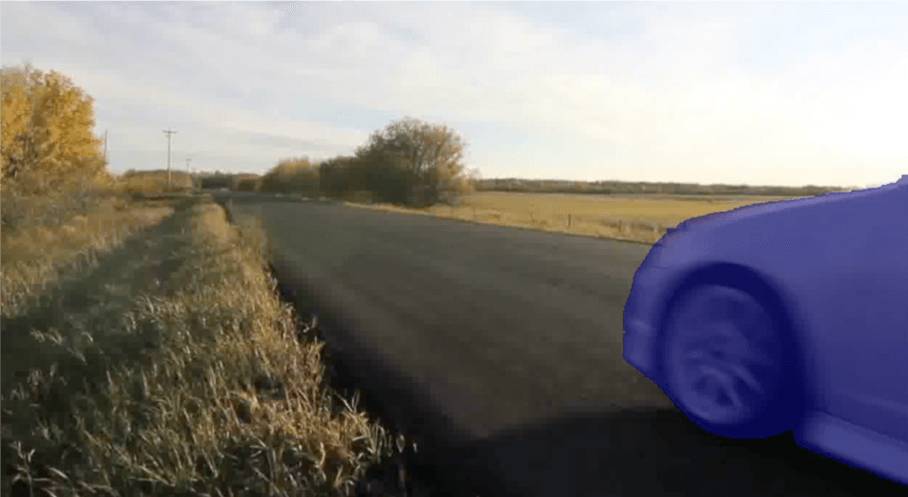} \\
%\rule{0pt}{8ex}
\vspace{1mm}
\image{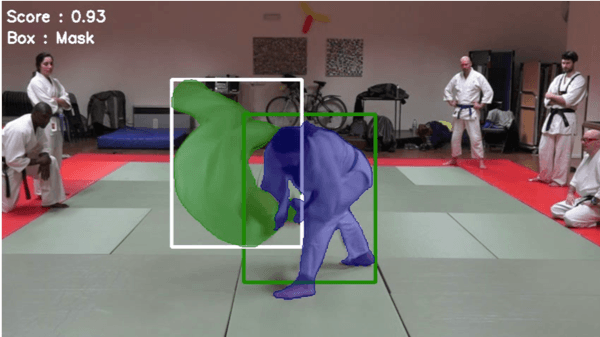} &
\image{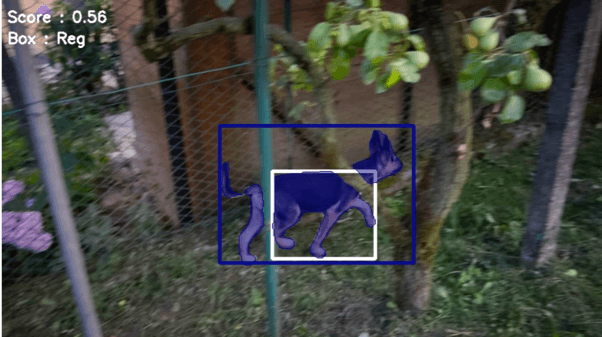} &
\image{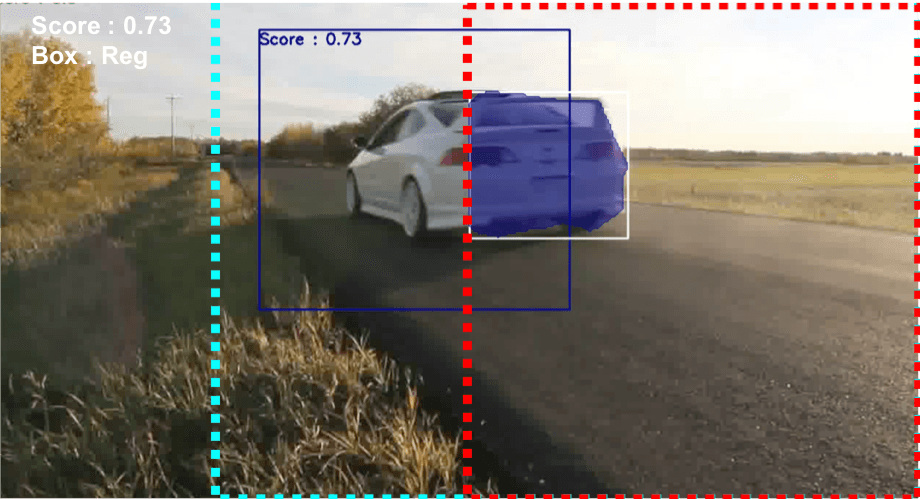} \\
%\rule{0pt}{8ex}
\vspace{1mm}
\image{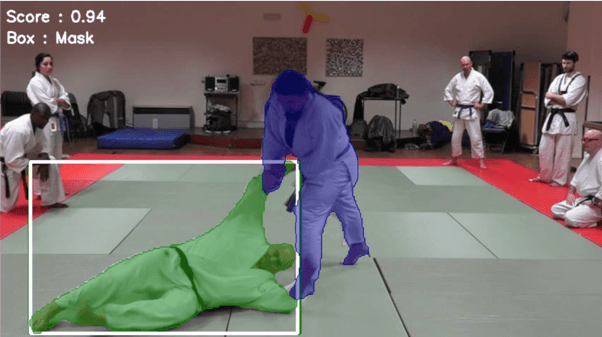} &
\image{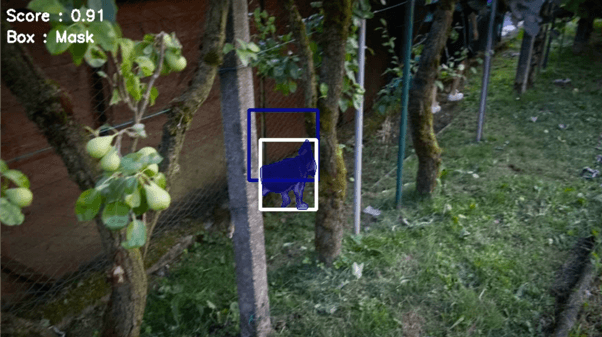} &
\image{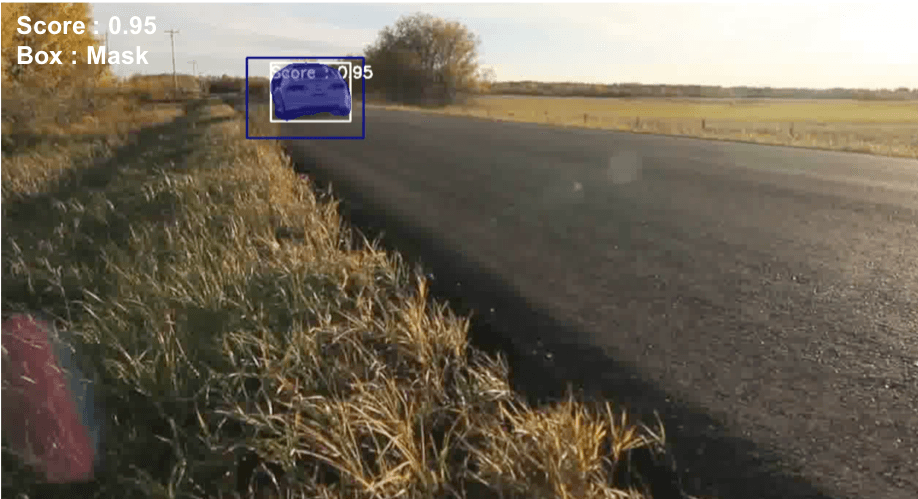} \\
\end{tabular}
\vspace{1mm}
\caption{Switches between the mask-box~(in white) and the regression-box~(in color). The first column shows that the mask-box is more robust to distractors. When the two players are twisted together~(second row), regression-box fails, and State Estimator chooses mask-box. The second column shows the regression-box provides a complete representation when the object is truncated or partially occluded. The third column shows that the regression-box can retrieve the target object in case of fast motion. The dotted line in {\color{cyan}cyan} represents the search region of the similarity encoder; the one in {\color{red}red} indicates the input region of the saliency encoder. }
\label{fig:state estimator}
\vspace{-3mm}
\end{figure}

\begin{figure}[H]
\centering
\includegraphics[width=\linewidth]{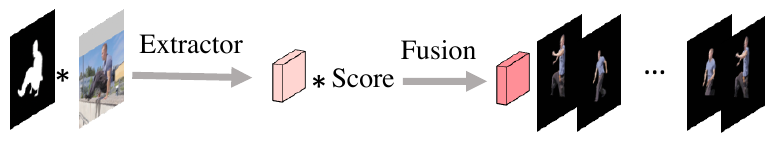}
\caption{Updating process of Global Modeling Loop. }
\label{fig:updating}
\end{figure}

With the above analysis, during inference, we pick mask-box for normal states to produce more accurate locations, while we choose regression-box for abnormal states to get more robust predictions.  Fig.~\ref{fig:state estimator} demonstrates some examples for strategy switching.
If we use mask-box for all frames, our model will lose track of the target when some abnormal states occur, otherwise if we keep using regression-boxes, we would get less accurate location predictions when the target is well-presented, or there are distractors in the background. Therefore, switching between these two strategies enables our model to make self-adaptions in different states and make our tracking process more accurate and stable.\\
%\indent After getting the box, we crop a search image of $303\times303$ on the next frame for the similarity encoder. And we crop a region that corresponds to a $127\times127$ area on original images and zoom it to $257\times257$ pixels for the saliency encoder. The scaling rule follows \cite{siamfc++}. \\
%\begin{comment}
%\end{comment}
\indent\textbf{\textit{ Global Modeling Loop}} : Global Modeling Loop updates a global feature for the target object dynamically, and uses this global feature to enhance the process of segmentation. As demonstrated in Fig.~\ref{fig:updating}, after predicting the binary mask for frame \textit{T} of target tracklet , we filter the background via element-wise multiplication. Then we feed the background-filtered image to a feature extractor~(shrinked ResNet-50) to get a neat target feature. Consider that all background-filtered frames share the same instance-level content, in spite that the appearance of the target object could change violently through the video flow. We fuse the high-level features of each background-filtered frame step by step to updates a robust global representation.  As Eq.~\ref{eq:global enhancer},  $\mathcal G $ denotes the global representation, and $\mathcal F $ denotes the high-level feature of the background-filtered image. $\mu $ denotes a hyper-parameter for step length that we set 0.5. Consider that if the target is occluded,  disappeared, or poorly segmented, the extracted feature would be useless or even harmful for the global representation. Therefore, we score the high-level feature of each frame with the state score $\mathcal S_{state}$ produced by State Estimator, thus alleviates adverse effects caused by abnormal situations or low-quality masks.
\begin{equation}
\mathcal G_t = (1- \mathcal S_{state} \cdot \mu) \cdot  \mathcal G_{t-1} +  \mathcal S_{state} \cdot \mu \cdot \mathcal F_t \label{eq:global enhancer}
\end{equation}

In this way, Global Modeling Loop updates a global feature that is robust for visual variants over time.  In return, we use this global feature to enhance the high-level representation of Joint Segmentation Network. This feedback loop makes our target representation more holistic and robust for long video sequences. \\

%------------------------------------------------------------------------
\section{Experiments}

\subsection{Network Training} 
The whole training process consists of two stages.  At the first stage, we train the similarity encoder and the regression head together on object tracking datasets~\cite{coco,deng2009imagenet,fan2019lasot,huang2018got,real2017youtubebb}. The training strategy follows 
SiamFC++~\cite{siamfc++}.
Then, we train the whole pipeline with the weight of the similarity encoder and the regression head frozen. The backbone of the saliency encoder and the feature extractor in Global  Modeling Loop are pretrained on ImageNet~\cite{deng2009imagenet}. For training data, we adopt COCO~\cite{coco}, DAVIS2017~\cite{davis20172017} training set~(60 videos) and the YouTube-VOS~\cite{xu2018youtube} training set~(3471 videos). We apply a cross-entropy loss on the predict binary mask of stride 4, and we also add auxiliary losses on the output feature of stride 8~(with weight 0.5) and stride 16~(with weight 0.3). 
We use SGD optimizer with momentum 0.9, set batch size as 16, and train our network on 8 GPUs with synchronized batch normalization. 
The training process takes about 8 hours, with 20 epochs. For each epoch, we select 160 000 images randomly. The first two epochs are a warm-up stage in which the learning rate increases linearly from $10^{-5}$ to $10^{-2}$. In the last 18 epochs, we apply a cosine annealing learning rate.

For each iteration, we randomly choose one target image and one search image from the same video sequence. The saliency encoder takes the cropped search image as input, while Global Modeling Loop picks the cropped target image. We use the ground truth mask to filter the background of the target image to train the extractor of Global Modeling Loop.

\subsection{Ablation Studies}
In Table \ref{tab:Ablation1}, we perform extensive ablations on DAVIS2017 validation dataset. We upgrade our model step by step from the most naive baseline to the full-version SAT to verify the effectiveness of each principal component. Then, we also explore the upper-bound of our method. \\
\indent\textbf{Naive Seg Baseline }: Our work starts from a naive segmentation baseline. We deal with each target as a tracklet, and we combine the saliency encoder and the decoder together to build a naive-segmentation network. For each video frame, we generate a min-max bounding box according to the predicted binary mask and crop a $257\times257$ search region for the next frame. This version performs weakly with only 48.1\% $\mathcal{J \& F}$ mean.  When the target object is truncated, occluded, or run out of the search region, the min-max bounding box generated by the predicted mask cannot locate the target object,  which causes target lost for successive frames.    \\
\indent\textbf{Track-Seg Baseline}: To tackle the problem of losing track.
We combine a siamese tracker~(SiamFC++\cite{siamfc++}) and the naive-segmentation network together. We use the siamese tracker to predict the target location and use the naive-segmentation network to produce a binary mask. This version gains excellent improvement compared with the naive baseline. However, it is still not able to deal with large pose/scale variations, and the segmentation accuracy is heavily constrained by the tracking quality.\\
\indent\textbf{Correlated Feature}: In order to obtain a more discriminative target representation, we introduce the correlated feature of similarity encoder to enhance the naive-segmentation network. The correlated feature contains appearance similarity, which brings 2.3\% improvement. \\
\indent\textbf{Global Modeling Loop }: For a more robust target representation over long sequences. We design Global modeling loop, which brings a significant improvement of 4.8\%. The effectiveness of the mask filter and state score weight is shown in the second part of Table \ref{tab:Ablation1}. Experiment results indicate that our idea of constructing a global representation is effective. Compared with only using the first frame or first frame + previous frame, the global representation brings $2.6\%$ and $1.2\%$ improvement respectively. We notice that the state score weight is also essential for updating the global representation, which improves the result by $1.2\%$.     
The effect of Global Modeling Loop is guaranteed by the mask filter, which brings a 5.6\% improvement. We find that the version without mask filter and the version which concatenates mask filter with images both bring adverse effects. We analyze that foreground objects of different frames share the same high-level semantic representation despite pose or scale changes, while the background keeps changing through the whole video. Therefore, foreground features of different frames are complementary for each other, while background features are not additive. Hence, an explicit process of background filtering is necessary. \\
\begin{table}[H]
\centering
\begin{threeparttable}
\begin{tabular}{lllll}
\toprule
Version & CF & GM & CS  & $\mathcal{J \& F}$ \\
%\midrule
\midrule
Naive Seg & & & &48.1    \\
Track-Seg & & & &61.6~({\color{red}+13.5})	 \\
Track-Seg &$\checkmark$ & & &63.9~({\color{red}+2.3})  \\
Track-Seg &$\checkmark$ &$\checkmark$ & &68.7~({\color{red}+4.8})	 \\
\textbf{Track-Seg~(SAT)} &$\checkmark$ &$\checkmark$ &$\checkmark$ &72.3~({\color{red}+3.6})	\\
\midrule
first + previous frame &$\checkmark$ & &$\checkmark$ &71.1~({\color{green}-1.2})\\
first frame only    &$\checkmark$ & &$\checkmark$ & 69.7~({\color{green}-2.6})\\
no Score Weight &$\checkmark$ & &$\checkmark$ &71.1~({\color{green}-1.2})\\
no Mask Filter &$\checkmark$ & &$\checkmark$ &66.7~({\color{green}-5.6}) \\
concat Mask &$\checkmark$ & &$\checkmark$  &66.5~({\color{green}-5.8}) \\
\midrule
Track-Seg   &$\checkmark$ & &$\checkmark$  &65.9~({\color{green}-6.4})	\\
Track-Seg      & &$\checkmark$ &$\checkmark$           &68.0~({\color{green}-4.3})	 \\
Naive Seg   &$\checkmark$ &$\checkmark$ &   &60.1~({\color{green}-12.2}) \\
\bottomrule
\end{tabular}
%\begin{tablenotes}
%\footnotesize
%\end{tablenotes}
\end{threeparttable}
\caption{Ablation studies for each component on DAVIS2017-Val dataset. \textbf{CF} denotes Correlated Feature. \textbf{GM} denotes Global Modeling Loop. \textbf{CS} denotes Cropping Strategy Loop. }
\label{tab:Ablation1}%添加标题 设置标签
\end{table}
\textbf{Cropping Strategy Loop}: In order to maintain a more stable tracklet. We construct Cropping Strategy Loop,  which switches the bounding box generation strategy according to the local state. This feedback loop brings a 3.6\% improvement. More importantly, the switching mechanism weakens the dependency for either tracking results or segmentation results, which enables us to use small backbones for each branch. 

We also analysis the switching mechanism by countering the usage rate of each strategy. On DAVIS2017-Val dataset, there are 30 sequences and 3923 frames in total. State Estimator judges 2876~(74\%) frames as normal states 1047~(26\%) as abnormal states. This statistic result agrees with our design intention that we use mask-box for the majority frames of normal states and regression-box for small numbers of abnormal situations.

\textbf{Upper-Bound Analysis}: As shown in Tab. \ref{tab:upper}, we explore the upper-bound of our pipeline by maximizing the effect of our two loops. 
For a clean global representation, we use the ground truth mask to filter the background of each frame, and this brings 1.7\% improvement. For an accurate bounding box for search region cropping, we use the ground truth mask to generate minimal bounding box, which brings 1.8\% improvement. In the ideal condition, the two loops make 5.2\% improvement together. Therefore, constructing a robust global representation and maintaining a stable tracklet are two topics that worth further study. 

%upper-bound
\begin{table}[H]
\centering
\begin{threeparttable}
\begin{tabular}{cccc}
\toprule
 & Mask Filter~(GT)  & Box~(GT)    & $\mathcal{J \& F}$ \\
%\midrule
\midrule
SAT &  &  & 72.3	\\
\midrule
SAT &$\checkmark$ &  & 74.0~({\color{red}+1.7})	\\
SAT &  &$\checkmark$ & 74.1~({\color{red}+1.8})	\\
SAT &$\checkmark$ &$\checkmark$ & 77.5~({\color{red}+5.2})	\\
\bottomrule
\end{tabular}
%\begin{tablenotes}
%\footnotesize
%\end{tablenotes}
\end{threeparttable}
\caption{  Upper-Bound for our pipeline. Mask GT means using the ground truth mask to filter the background for global guidance. Box GT means using the ground truth bounding box to crop the search region for the next frame.}
\label{tab:upper}%添加标题 设置标签
\end{table}

\subsection{Comparison to state-of-the-art}
We evaluate our method on DAVIS2017-Val~\cite{davis20172017}, DAVIS2016-Val~\cite{davis2016benchmark} and YouTube-VOS\cite{xu2018youtube} datasets. Quantitative results demonstrate that our approach achieves promising performance for both accuracy and speed.

\textbf{DAVIS2017}: For the task of multi-object VOS, we predict a probability map for each target, then we concatenate them together, and apply a softmax aggregation to get the final result.
We compare SAT with state-of-the-art methods. 
%We do not compare  PReMVOS~\cite{luiten2018premvos}, BoltVOS~\cite{boltvos} and PTS~\cite{PTS} because they combine several very large network together and run in extremely low frame rates. 
For the evaluation metrics, $\mathcal{J \& F}$ evaluates the general quality of VOS result, $\mathcal{J}$ estimates the mask IOU, $\mathcal{F}$ describes the quality of contour. $\mathcal{J}_{\mathcal{D}}$ denotes the performance decay of $\mathcal{J}$ over time. FPS is measured for the time of every forward pass on a single RTX 2080Ti GPU.  

As shown in Tab. \ref{tab:DV17}, some newly proposed methods like FEELVOS~\cite{voigtlaender2019feelvos}, AGAME~\cite{agame} aim to make the balance between speed and accuracy but SAT gets a more promising result for both. %DOLF~\cite{dolf} applies an efficient online learning strategy and gets a slightly higher $\mathcal{J \& F}$, but SAT runs 2.5 times faster. 
SiamMask~\cite{siammaks} and RANet~\cite{wang2019ranet} also run at real-time speed, but their segmentation accuracy is obviously worse than ours.
In general, SAT surpassed most of  newly proposed models for both accuracy and efficiency.

SAT gets the best running speed and contour quality while achieves the highest $\mathcal{J \& F}$ among newly proposed methods. Besides, SAT has the lowest performance decay $\mathcal{J}_{\mathcal{D}}$, which means our method is robust over time, and we would gain more advantages over others for long sequences. At the bottom row of Tab. \ref{tab:DV17},  We also develop a faster version with ResNet-18 backbone, which runs at 60 FPS with slightly lower prediction accuracy.\\   
\indent\textbf{YouTube-VOS}: We mainly compare our method with some fast and offline learning methods on YouTube-VOS benchmark. Tab. \ref{tab:youtube-vos} shows our method achieves competitive  performance and surpasses~\cite{xu2018youtube,RGMP,siammaks} for both seen and unseen categories.    \\
\indent\textbf{DAVIS2016}: Single object segmentation is a relatively simpler task. As shown in Tab.~\ref{tab:DV16}, online fine-tuning often brings huge promotion on DAVIS2016 while costs enormous computation. Hence, we mainly compare our method with some newly proposed offline models. SAT performs better than FEELVOS~\cite{voigtlaender2019feelvos}, AGAME~\cite{agame}, RGMP~\cite{RGMP} and SiamMask~\cite{siammaks}.\\
\indent\textbf{Computation Analysis}: Running speed can be influenced by the environment and hardware condition. For a fair comparison, we also counter the multiply-accumulate operations of several fast VOS models. As shown in Tab. \ref{tab:flops}, our method costs obvious fewer Gflops than others. The computation of CNNs is highly related to input resolution and backbone size. Each component of SAT is specially designed for efficiency. The similarity encoder has a large input of $303 \times 303$, so we pick Alexnet as the backbone. The saliency encoder takes $257 \times 257$ image as input, and we use a shrinked ResNet-50 backbone, in which we set the channel expansion rate as 1. 
Global Modeling Loop only cares about the high-level feature, so we resize the filtered images to $129 \times 129$.
In contrast, RANet~\cite{wang2019ranet} and AGAME~\cite{agame} use ResNet-101 backbone with $480 \times 864$ input size, which makes them computational expensive. SiamMask~\cite{siammaks} takes $255 \times 255$ images as input and uses a ResNet-50 backbone, and it replaces the stride-2 convolutions of the last two stages to stride-1, which helps to keep spatial information but brings more computation. Besides, SiamMask follows DeepMask\cite{deepmask} to apply a pixel-wise mask representation, which entails much computation.

%dv2017
\begin{table}[h]
\centering
\small
\begin{tabular}{cp{0.2cm}ccccc}
\toprule
Method & OL  & $\mathcal{J \& F}$ &  $\mathcal{J}_{\mathcal{M\uparrow}}$  &  $\mathcal{J}_{\mathcal{D\downarrow}}$  & $\mathcal{F}_{\mathcal{M\uparrow}}$  & FPS \\
\midrule
%PReMVOS\cite{luiten2018premvos}	&$\checkmark$ &{\color{red}77.8}	&{\color{red}73.9} &16.2  &	{\color{red}81.7} &	0.01\\
PReMVOS\cite{luiten2018premvos} &$\checkmark$ &77.8 &73.9 &16.2 &81.7 &0.01\\
%DOLF\cite{dolf}	&$\checkmark$	&73.4	&71.5 &-	&75.5	&14.6\\
%DOLF-f\cite{dolf}	&$\checkmark$	&70.9	&68.4 &-	&73.3 	&25.3\\
OSVOS-s\cite{osvos-s}	& $\checkmark$	&68.0	&64.7 &15.1	&71.3 	&0.22\\
OnAVOS\cite{onavos}	& $\checkmark$	&67.9	&64.5 &27.9	&71.2 	&0.08\\
CINM\cite{cinm}	& $\checkmark$	&67.5	&64.5 &24.6	&70.5 	& 0.01\\ 
Dyenet\cite{dyenet}	& $\checkmark$	&69.1	&67.3 &-	&71.0 	&2.4\\
OSVOS\cite{osvos}	& $\checkmark$	&60.3	&56.7 &26.1	&63.9 	&0.22\\
\midrule
*STM\cite{stm} &$\times$ &{\color{red}\textbf{81.8}} &{\color{red}\textbf{79.2}} &- &{\color{red}\textbf{84.3}} &6.25\\

FEELVOS \cite{voigtlaender2019feelvos} &$\times$ &71.5 &69.1 &17.5 &74.0
&2.2 \\
AGAME\cite{agame}	&$\times$	&70.0	 &67.2 &{\color{blue}\textbf{14.0}}	&72.7 	&14.3\\
RGMP\cite{RGMP}	&$\times$	&66.7	&64.8 &18.9	&68.6 	&7.7\\
RANet\cite{wang2019ranet}	&$\times$	&65.7	&63.2 &18.6	&68.2 	&30\\
STCNN\cite{STCNN}	&$\times$	&61.7	&58.7 &-	&64.6 	&0.25\\
FAVOS\cite{FAVOS}	&$\times$	&58.2	&54.6 &14.4	&61.8 	&0.56\\
SiamMask\cite{siammaks}	&$\times$	&56.4	&54.3 &19.3	&58.5 &35\\

%\midrule
Ours	&$\times$	&{\color{blue}\textbf{72.3}}	&{\color{blue}\textbf{68.6}} &{\color{red}\textbf{13.6}}	&{\color{blue}\textbf{76.0}}	 &{\color{blue}\textbf{39}}\\

Ours-Fast	&$\times$	&69.5	&65.4 &16.6	&73.6 &{\color{red} \textbf{60}}\\

\bottomrule
\end{tabular}
\caption{Quantitative results on DAVIS2017 validation set. OL denotes online fine-tuning. FPS denotes frame per second. The best two results among offline methods are marked in {\color{red}red} and {\color{blue}blue} respectively. 
*: STM requires more training data and longer training time than other works.} 
\label{tab:DV17}%添加标题 设置标签
\end{table}

%flops 
\begin{table}[h]
\centering 
\small
\tabcolsep=2.3pt
\begin{tabular}{c|ccccc}
\toprule
Method & Ours-f & Ours & SiamMask~\cite{siammaks} & RANet~\cite{wang2019ranet}  & AGAME~\cite{agame} \\
\midrule
Gflops  & $ \sim 12 $ 	& $ \sim 13 $  &$ \sim 16 $  & $>65$  & $>65$\\
\midrule
FPS     & 60 & 39 & 35 & 30 & 14.3 \\
\bottomrule
\end{tabular}
\caption{Computation analysis for some fast VOS models, Gflops counters multiply-accumulate operations. Ours-f denotes the fast version SAT with a Alexnet backbone.  }
\label{tab:flops}%添加标题
\end{table}

%ytb-vos
\begin{table}[h]
\centering 
\small
\begin{tabular}{ccccccc}
\toprule
Method  & OL & $\mathcal{G}  $  &  $\mathcal{J}s$ & $\mathcal{J}u$  &  $\mathcal{F}s$ & $\mathcal{F}u$ \\
\midrule
%DOLF\cite{dolf}   &71.0 &71.6 &65.0 &74.7 &72.5 &14.6\\
PreMVOS\cite{luiten2018premvos}  &$\checkmark$ &66.9 &71.4  &56.5 &75.9 &63.7\\
OSVOS\cite{osvos}   &$\checkmark$ &58.8 &59.8  &54.2 &60.5  &60.7  \\
OnAVOS\cite{onavos}	 &$\checkmark$ &55.2 &60.1 &46.1 &62.7 &51.4  \\
\midrule
%AGAME\cite{agame} &66.0 &66.9 &61.2 &68.6 &67.3 &14.3\\
*STM\cite{stm} &$\times$ &{\color{red}\textbf{79.4}} &{\color{red}\textbf{79.7}} &{\color{red}\textbf{84.2}} &{\color{red}\textbf{72.8}} &{\color{red}\textbf{80.9}} \\
S2S\cite{xu2018youtube}  &$\times$ &57.6 &66.7 &48.2  &- &- \\

RGMP \cite{RGMP}   &$\times$  &53.8 &59.5 &45.2 &- &-  \\
SiamMask\cite{siammaks}	 &$\times$ &52.8	&60.2 &45.1 &58.2  &47.7 \\

Ours &$\times$	&{\color{blue}\textbf{63.6}}	&{\color{blue}\textbf{67.1}} &{\color{blue}\textbf{55.3}} &{\color{blue}\textbf{70.2}}	 &{\color{blue}\textbf{61.7}}  \\

\bottomrule
\end{tabular}
\caption{Quantitative results on Youtube-VOS benckmark. OL denotes online fine-tuning. The subscript $s$ denotes seen categories while $u$ denotes unseen categories.  The best two results among offline methods are marked in {\color{red}red} and {\color{blue}blue} respectively. *: STM requires more training data and longer training time than other works.
} 
\label{tab:youtube-vos}%添加标题
\end{table}

%dv2016
\begin{table}[h]
\centering
\begin{tabular}{cccccc}
\toprule
%Method & Ol  & $\mathcal{J \& F}$ &  $\mathcal{J}$ & $\mathcal{F}$  & FPS \\
Method & OL  & $\mathcal{J \& F}$ &  $\mathcal{J}_{\mathcal{M\uparrow}}$  & $\mathcal{F}_{\mathcal{M\uparrow}}$  & FPS \\
\midrule
RANet+\cite{wang2019ranet}	&$\checkmark$	&87.1	&86.6	&87.6	&0.25\\
PReMVOS\cite{luiten2018premvos}	&$\checkmark$ &86.8 &84.9 &88.6&	0.01\\
%DOLF\cite{dolf}	&$\checkmark$   &84.8 &85.0 &84.5	&14.6\\
%OSVOS-s\cite{osvos-s}	&$\checkmark$	&86.5 &85.6 &87.5	&0.22\\
%OnAVOS\cite{onavos}	&$\checkmark$	&85.5 &86.1 &84.9	&0.08\\
%DOLF-f\cite{dolf}	&$\checkmark$	&82.5 &82.7 &82.4	&25.3\\
OSVOS\cite{osvos}	&$\checkmark$	&80.2	&79.8	&80.6	&0.22\\
\midrule
%RANet\cite{wang2019ranet}	&$\times$	&{\color{red}\textbf{85.5}}	&{\color{red}\textbf{85.5}}	&{\color{red}\textbf{85.4}}	&30\\
*STM\cite{stm} &$\times$  &{\color{red}\textbf{89.3}} &{\color{red}\textbf{88.7}} &{\color{red}\textbf{89.9}} &6.25 \\ 
RGMP\cite{RGMP}	&$\times$	&81.8 &81.5 &82.0	&7.7\\
AGAME\cite{agame}	&$\times$	&- &82.0 &-	&14.3\\
FEELVOS\cite{voigtlaender2019feelvos}	&$\times$	&81.7 &81.1 &82.2	&2.2\\
FAVOS\cite{FAVOS}	&$\times$	&80.8 &82.4 &79.5	&0.56\\
SiamMask\cite{siammaks}	&$\times$	&69.8	&71.7	&67.8	&{\color{ blue}\textbf{35}}\\
%\midrule
Ours	&$\times$	&{\color{blue}\textbf{83.1}}	&{\color{blue}\textbf{82.6}}	&{\color{blue}\textbf{83.6}}	& {\color{red} \textbf{39}}\\

\bottomrule
\end{tabular}
\caption{Quantitative results on DAVIS2016 validation set. OL denotes online fine-tuning. FPS denotes frame per second.  The best two results among offline methods are marked in {\color{red}red} and {\color{blue}blue} respectively.  *: STM requires more training data and longer training time than other works.
}
%The best results among offline methods are marked in {\color{red}red}, while the second-best results in {\color{blue}blue}.}
\label{tab:DV16}%添加标题 设置标签
\end{table}

%demo
\begin{figure*}[h]
\newcommand{\image}{\includegraphics[width=0.40\columnwidth]}
\centering 
\tabcolsep=0.05cm
\renewcommand{\arraystretch}{0.12}
\begin{tabular}{ccccc}
 %{\footnotesize Distractor} & {\footnotesize Truncation} & {\footnotesize Fast Motion} \\
\vspace{1mm}
\image{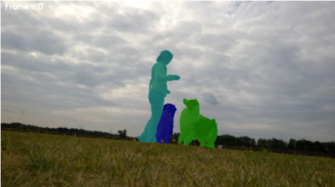} &
\image{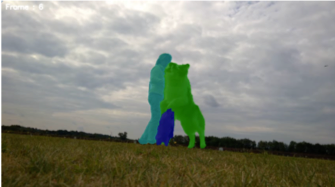} &
\image{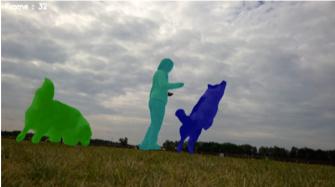} &
\image{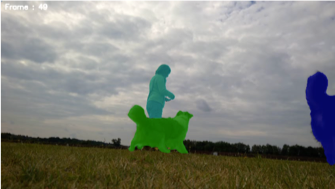} &
\image{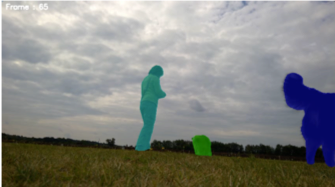} \\
\vspace{1mm}
\image{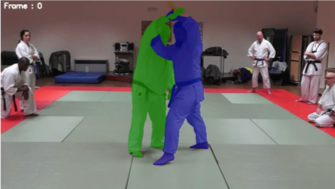} &
\image{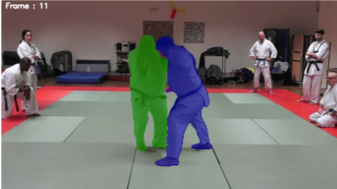} &
\image{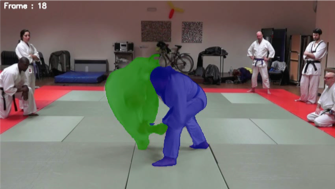} &
\image{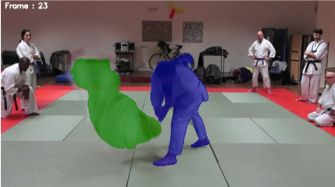} &
\image{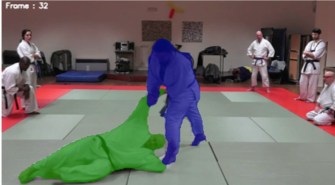} \\
%\rule{0pt}{8ex}
\vspace{1mm}
\image{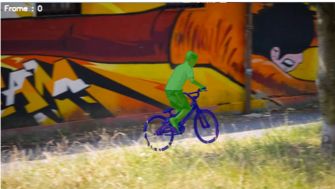} &
\image{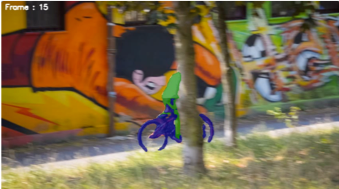} &
\image{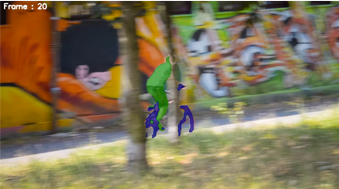} &
\image{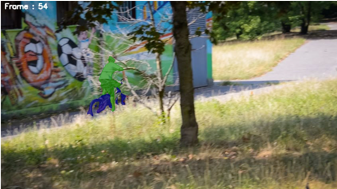} &
\image{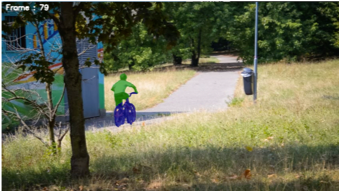} \\
\vspace{1mm}
\image{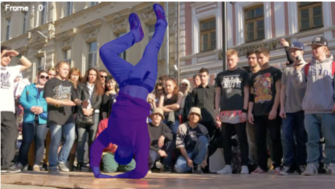} &
\image{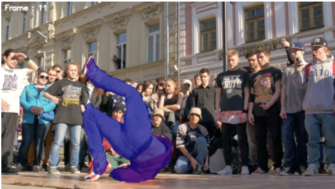} &
\image{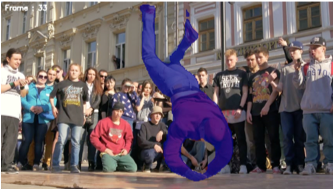} &
\image{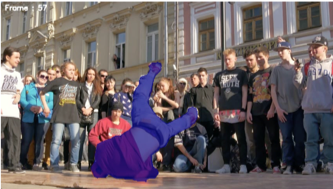} &
\image{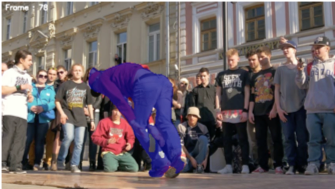} \\
\begin{comment}
\vspace{1mm}
\image{Figure_dir/davis/5_1.png} &
\image{Figure_dir/davis/5_2.png} &
\image{Figure_dir/davis/5_3.png} &
\image{Figure_dir/davis/5_4.png} &
\image{Figure_dir/davis/5_5.png} \\
\end{comment}

\end{tabular}
\vspace{1mm}
\caption{Qualitative results of SAT on DAVIS Benchmark. }
\label{fig:demos}
\vspace{-3mm}
\end{figure*}

\subsection{Qualitative result} 
Fig.~\ref{fig:demos} shows the qualitative result of our method on DAVIS benchmarks. SAT can produce robust and accurate segmentation results even in complicated scenes. The first three rows show that SAT is robust for distractors, motion blur and occlusion. The last row shows that SAT is robust for tremendous pose variant. 

\section{Conclusion}
In this paper, we present State-Aware Tracker~(SAT), which achieves promising performance with high efficiency for the task of semi-supervised video object segmentation. SAT takes each target object as a tracklet to perform VOS more efficiently. With an Estimation-Feedback mechanism, SAT can get awareness for the current state and make self-adaptation to reach stable and robust performance. Our methods achieves competitive performance on several VOS benchmarks with a decent speed-accuracy trade-off.

\small{\textbf{Acknowledgements}: This paper is supported by the National key R\&D plan of the Ministry of science and technology (Project Name: “Grid function expansion technology and equipment for community risk prevention”, Project No. 2018YFC0809704)}

{\small
\bibliographystyle{ieee_fullname}
\bibliography{egbib}
}

\end{document}